\pdfoutput=1
\documentclass[sigconf]{acmart}
\usepackage{booktabs} % For prettier tables
\usepackage{siunitx} % For aligning numbers by decimal point
\usepackage{tabularx} % For more flexible table layouts
\newcolumntype{Y}{>{\centering\arraybackslash}X} % Centering content in X columns
\usepackage{comment}
\usepackage{graphicx}
\usepackage{svg}
\usepackage{amsmath}
\usepackage{amsfonts}
\AtBeginDocument{%
  \providecommand\BibTeX{{%
    \normalfont B\kern-0.5em{\scshape i\kern-0.25em b}\kern-0.8em\TeX}}}

\RequirePackage[skins]{tcolorbox}
\newtcolorbox{custombox}[1]{
	colback=gray!10,
	colframe=gray!70,
	left=1mm,
	right=1mm,
	top=1mm,
	bottom=1mm,
	fonttitle=\bfseries,
	arc=0mm,
	leftrule=0mm,
	rightrule=0mm,
	toprule=0mm,
	bottomrule=0mm,
	notitle,
	before=\noindent,
	before upper={#1  },
}

\usepackage{etoolbox}
\usepackage{booktabs}
\usepackage{wrapfig}
\copyrightyear{2024}
\acmYear{2024}
\setcopyright{acmlicensed}\acmConference[ACE 2025]{}
\acmBooktitle{}
\acmBooktitle{Australian Computing Education Conference (ACE 2025), January 29-February 2, 2024, Brisbane, Australia}
\acmPrice{15.00}
\acmDOI{}
\acmISBN{}

\begin{document}

%\title{Investigating the Capabilities of Large Multimodal Models to Solve Visually Diverse Graph and Tree Problems}

%\title{Large Multimodal Model Performance on Graph and Tree Tasks}
\title[Large Multimodal Models Can Solve Diverse Graph and Tree Vision Tasks]{Large Multimodal Models Can Solve Diverse Graph and Tree Vision Tasks}
\title[Seeing the Forest and the Trees]{Seeing the Forest and the Trees: Solving Visual Graph and Tree Based Data Structure Problems Using Large Multimodal Models}


%\author{Irene Y. Hou, Owen Man, Kenneth Angelikas, Sophie Metille, Stephen MacNeil}
%\affiliation{\institution{Temple University}
%             \department{Department of Computer and Information Sciences}}

% \author{Stephen MacNeil} % Steve 
% \email{smacneil@ucsd.edu}
% \affiliation{
%     \department{Department of Computer and Information Sciences}
%     \institution{Temple University}
%     \city{Philadelphia}
%     \state{Pennsylvania}
%     \country{USA}
% }

\author{Sebastian Gutierrez}
\affiliation{
  \institution{Temple University}
  \streetaddress{1801 N Broad St}
  \city{Philadelphia}
  \state{PA}
  \country{USA}
  \postcode{19122}
}
\email{guts@temple.edu}
\orcid{0009-0005-4844-692X}

\author{Irene Hou}
\affiliation{
  \institution{Temple University}
  \streetaddress{1801 N Broad St}
  \city{Philadelphia}
  \state{PA}
  \country{USA}
  \postcode{19122}
}
\email{irene.hou@temple.edu}
\orcid{0009-0008-0511-7685}

\author{Jihye Lee}
\affiliation{
  \institution{Temple University}
  \streetaddress{1801 N Broad St}
  \city{Philadelphia}
  \state{PA}
  \country{USA}
  \postcode{19122}
}
\email{jihye.lee0003@temple.edu}
\orcid{0009-0008-0511-7685}

\author{Kenneth Angelikas}
\affiliation{
  \institution{Temple University}
  \streetaddress{1801 N Broad St}
  \city{Philadelphia}
  \state{PA}
  \country{USA}
  \postcode{19122}
}
\email{kenneth.angelikas@temple.edu}
\orcid{0009-0001-1597-3513}

\author{Owen Man}
\affiliation{
  \institution{Temple University}
  \streetaddress{1801 N Broad St}
  \city{Philadelphia}
  \state{PA}
  \country{USA}
  \postcode{19122}
}
\email{owen.man@temple.edu} 
\orcid{0009-0003-0527-1395}

\author{Sophia Mettille}
\affiliation{
  \institution{Temple University}
  \streetaddress{1801 N Broad St}
  \city{Philadelphia}
  \state{PA}
  \country{USA}
  \postcode{19122}
}
\email{sophia.mettille@temple.edu}
\orcid{0009-0009-9707-6311}

\author{James Prather}
\orcid{0000-0003-2807-6042}
\affiliation{
  \institution{Abilene Christian University}
  \city{Abilene}
  \state{TX}
  \country{USA}
}
\email{james.prather@acu.edu}

\author{Paul Denny}
\orcid{0000-0002-5150-9806}
\affiliation{
  \institution{The University of Auckland}
  \city{Auckland}
  \country{New Zealand}
}
\email{paul@cs.auckland.ac.nz}

\author{Stephen MacNeil}
\affiliation{
  \institution{Temple University}
  \streetaddress{1801 N. Broad St}
  \city{Philadelphia}
  \state{PA}
  \country{USA}
  \postcode{19122}
}
\email{stephen.macneil@temple.edu}
\orcid{0000-0003-2781-6619}

\newcommand{\td}[1]{{\color{black} #1}}
\newcommand{\fb}[1]{{\color{black} #1}}

\renewcommand{\shortauthors}{Gutierrez, et al.}


\begin{abstract}

\fb{%Recent advancements in the development of large multi-modal models (LMMs) demonstrate capable performance in zero-shot text prompting. These advancements, coupled with evaluations of conjunctive text-vision modalities, shows that text-in-image abilities meet similar standards. With that being said, transformer-based it is unclear how well they handle visual problems like graphs and trees, 4) we ran a study, 5) results..., 6) implications

% 1) LMMs really good at text 
% 2) They are also really good at image extraction and analysis but...
% 3) are they good at full on vision tasks?
% 4) Our goal is to understand whether or not LMM are capable of visual question answering tasks 
% 5) our study provides a corpus of 10,000 ... for further use for testing the capabilities of LMM, a comprhensive analysis on the performance of three diffrent vision models, and a comprhensive method on top of eveythign built for testing vision capabilities of LMM.
% 6) our study shows RESULTS (needs to be framed for followign statement)
% 7) The implications of these results go further than just vision capabilities but we aim to put the final nail in the coffin of assesment. With these models rapid increase in perfromance moth over month we must understand that educators can no longer just keep finding work arounds for these models but instead shift from these extrinstic motivators to intrinstic motivatiors. 

Recent advancements in generative AI systems have raised concerns about academic integrity among educators. Beyond excelling at solving programming problems and text-based multiple-choice questions, recent research has also found that large multimodal models (LMMs) can solve Parsons problems based only on an image. However, such problems are still inherently text-based and rely on the capabilities of the models to convert the images of code blocks to their corresponding text. In this paper, we further investigate the capabilities of LMMs to solve graph and tree data structure problems based only on images. To achieve this, we computationally construct and evaluate a novel benchmark dataset comprising \textbf{9,072} samples of diverse graph and tree data structure tasks to assess the performance of the GPT-4o, GPT-4 with Vision (GPT-4V), Gemini 1.5 Pro, Gemini 1.5 Flash, Gemini 1.0 Pro Vision, and Claude 3 model families. GPT--4o and Gemini 1.5 Flash performed best on trees and graphs respectively. GPT-4o achieved 87.6\% accuracy on tree samples, while Gemini 1.5 Flash, achieved 56.2\% accuracy on graph samples. Our findings highlight the influence of structural and visual variations on model performance. This research not only introduces an LMM benchmark to facilitate replication and further exploration but also underscores the potential of LMMs in solving complex computing problems, with important implications for pedagogy and assessment practices.

}
\end{abstract}


% Here is the table for all ACM CSS concepts https://dl.acm.org/ccs and I find Human-centered computing -> Visualization, Collaborative and social computing; Applied computing





    %\centering
    %\includegraphics[width=\textwidth]{mosaic_8x2.png}
    %\caption{The many parametrically generated graphs and trees used for this study.}

\maketitle

\section{Introduction}

% Frame for Introduction
% 1) 

% Motivating the problem that LLMs introduce around assessment

Generative AI is increasingly prevalent in computing education, with diverse applications that range from code generation~\cite{chen2021evaluating, barke2022grounded, denny2023conversing, puryear2022github, wermelinger2023using} and high-quality code explanations~\cite{macneil2022generating, bernstein2024like, bernstein2024analyzing, macneil2023experiences, leinonen2023comparing, wermelinger2023using} to debugging code~\cite{macneil2023decoding, ouh2023chatgpt}. Consequently, many students now report using them as a primary help resource~\cite{hou2023effects, prather2023robots, zastudil2023generative}. However, the capabilities of generative AI extend beyond mere coding support; students can also leveraging these technologies to answer multiple-choice questions~\cite{savelka2023generative, savelka2023thrilled} and solve text-based programming problems~\cite{reeves2023evaluating, hou2024more, finnie-ansley2023my}. %In recent events, newer models have promised to be multimodal, real-time personal teaching assistants~\footnote{https://www.khanmigo.ai/}. 
The emergence of multimodal models has further transformed this landscape by enabling the comprehension and generation of visual data, thereby enhancing the models' ability to engage with complex educational tasks. While this advancement presents new opportunities for learning, it simultaneously raises significant concerns among researchers and educators regarding assessment integrity and the potential for misuse in academic settings~\cite{macneil2023implications, prather2023transformed, prather2023robots, lau2023from, zastudil2023generative}.

%Recently, multimodal models have been developed with the ability to understand and generate visual data. This has raised serious concerns from researchers and practitioners in computing education related to  assessment and potential for misuse~\cite{macneil2023implications, prather2023transformed, prather2023robots, lau2023from, zastudil2023generative}.

%Computing education researchers have recently expressed both concern and excitement about the ways that generative models may affect the computing education landscape~\cite{macneil2023implications, macneil2022automatically, prather2023transformed, lau2023from, zastudil2023generative}. In all of these cases, assessment has been a central topic of discussion. %even display general emergent abilities that require further probing~\cite{wei2022emergent, bubeck2023sparks}. 
% Work on parsons problems and visual assignments 

To deter unsanctioned use of large language models (LLMs) among students, researchers have proposed solutions such as human-proctored exams~\cite{zastudil2023generative, Joshi2023ChatGPT, rudolph2023chatgpt, prather2023robots, susnjak2022chatgpt} and visual-based questions that challenge the text-only modalities of LLMs~\cite{denny2023promptly, ouh2023chatgpt, cipriano2024picture}. However, human-proctored exams come with inherent limitations, including biases in test-taking and scalability challenges that can hinder their effectiveness~\cite{davis2013racial, holmes2021bad, xie2021domain, conijn2022fear}. Although AI-proctored systems offer a potential solution to scalability issues for online exams, they introduce a range of new concerns, such as breaches of privacy and autonomy, barriers to accessibility, and the risk of negative biases~\cite{gonzalez2020implementation, coghlan2021good, gudino2021remote, cahapay2021problems}. While representing programming problems visually—through diagrams or images—has historically limited LLMs' problem-solving capabilities~\cite{denny2023promptly, ouh2023chatgpt}, the advent of multimodal models that can process both images and text poses a significant challenge to this approach. Recent studies, such as those by Hou et al., have demonstrated these models' remarkable performance on visual programming problems, including Parsons problems~\cite{hou2024more}. This development underscores the urgent need for further investigation into the capabilities of multimodal models in visual problem-solving contexts. As these models become increasingly proficient in handling complex computational tasks, the potential for academic dishonesty grows, complicating traditional assessment strategies.% and raising critical questions about the integrity of student evaluations.

%Thus far, in CS1 contexts, these models appear to dominate. But as students progress in their coursework, the programming problems they face become increasingly abstract, with an emphasis on foundational concepts such as tree and graph data structures, which are often visually represented and evaluated in computing courses~\cite{joint2013computer}. 

This paper builds upon previous research regarding the vision capabilities of large multimodal models (LMMs) by systematically evaluating the performance of several advanced models, including GPT-4o, GPT-4V, Gemini 1.5 Flash, Gemini 1.5 Pro, Gemini 1.0 Pro Vision, and the Claude 3 model family. To mitigate data leakage, we computationally constructed a benchmark dataset comprising 9,072 diverse graph and tree data structure problems, designed to assess how these models handle variations in both \textit{aesthetic features (such as edge width and node color)} and \textit{structural characteristics (including density, layout, and the number of nodes and edges)}. Through this evaluation, we aim to gain insights into the strengths and limitations to inform computing education pedagogy and assessment.

%This paper extends previous work on LMM vision capabilities by evaluating the performance of multiple LMMs (GPT-4o~\footnote{https://openai.com/index/hello-gpt-4o/}, GPT-4V~\cite{openai2024gpt4}~\footnote{https://cdn.openai.com/papers/GPTV\_System\_Card.pdf}, Gemini 1.5 Flash ~\footnote{https://deepmind.google/technologies/gemini/flash/}, Gemini 1.5 Pro~\cite{reid2024gemini}, Gemini 1.0 Pro Vision~\cite{geminiteam2024gemini}, and the Claude 3 model family~\footnote{https://www-cdn.anthropic.com/de8ba9b01c9ab7cbabf5c33b80b7bbc618857627/Model\_Card\_Claude\_3.pdf}) on visually varied images of tree and graph structure tasks. We computationally developed and evaluated the performance of these models on a dataset of 9,072 graph and tree data structure problems that varied in aesthetic features (edge width, node color) and structural makeup (density, layout, number of nodes, and edges).

We investigate the following research questions: 

%We compared the large multimodal model (LMM) GPT-4V~\footnote{https://cdn.openai.com/papers/GPTV\_System\_Card.pdf} with Bard~\footnote{https://bard.google.com/}, a large language model (LLM), that uses Google Lens for text recognition. We find that LMMs, which have learned both pixel features (from images) and text features (from prompts) in the same embedding space, performed substantially better than Bard which uses a piecemeal approach.
%In this paper, we investigate the performance of two large multimodal vision models (i.e.: GPT-4V~\footnote{https://cdn.openai.com/papers/GPTV\_System\_Card.pdf} and Bard~\footnote{https://bard.google.com/}) 
%As a first step in understanding their capabilities to solve visual problems, We evaluated their performance on multiple Parsons problems across diverse visual representations. We created a dataset of Visual Parsons problems based on existing literature~\cite{reeves2023evaluating}, then represented each Parsons problem across diverse visual formats based on common Parsons Problem generation tools include JS-Parsons \cite{Sirkia2016ParsonVisulaization, helminen2012StudentsParson, Ihantola2011Parosn2D, Ihantola2013ProgrammingMobile}, Runestone~\cite{ericson2020Runestone, YeckehZaare2019SpacedTool, ericson2016IdentifyingDesign, ericson2015InteractiveEbook}, Epplets~\cite{kumar2018epplets, kumar2017effect}, and EvoParsons~\cite{bari2019evoparsons, gaspar2019lessons}. We then prompted GPT-4V and Bard with pre-engineered prompts to evaluate their ability to solve these images of visually diverse Parsons problems. 

\begin{enumerate}
    \item [\textbf{RQ 1}] How well do large multi-modal models perform operational and representational tasks on graph and tree data structures? %We show current state-of-the-art systems can achieve passing scores on certain sets of tasks while others remain a great challenge to models.
    \item [\textbf{RQ 2}] How do variations in the graph and tree's structural features influence the accuracy of large multimodal models? %We extract features to find the influence of structural properties like degrees, edges, and more.
    \item [\textbf{RQ 3}] How do variations in the graph and tree's aesthetic features influence the accuracy of large multimodal models? %We extract features to find the influence of aesthetic properties like node color and edge width.
\end{enumerate}

%RQ1: problem
%how effectively do...
%RQ2: structure 
%how do structural variations influence the...
%RQ3: visual representation
%how do visual representations influence...node placement/layout
%node number
%node values
%visual formats

Our results indicate that these multimodal models are capable of performing operational and representational tasks on graphs and trees to varying degrees, depending on structural and aesthetic variations. GPT-4o was a top-performer on tree-based problems with 87.6\% accuracy, but only 44.7\% on graph problems. This study highlights both the challenges that models face and the growing concerns for educators, particularly as LMMs are expected to continue improving. To facilitate replication and further research with new models, we also contribute an open-source repository for generating graph and tree problems. We conclude with a discussion on the implications for pedagogy and assessment practices. We offer the following contributions: 

\begin{itemize}
\item \textit{Empirical evaluation of multimodal models on data structure tasks:} We provide a comprehensive analysis of multiple multimodal models' performance on visually represented graph and tree problems, highlighting their varying abilities to handle different structural and aesthetic aspects. 
\item \textit{A benchmark dataset for assessing multimodal model capabilities:} We introduce a novel dataset of 9,072 data structures problems that vary across structural and visual attributes. 
\item \textit{An open-source tool for generating graph and tree problems:} We contribute a tool to create diverse graph and tree problems, supporting further research and the development of new benchmarks as multimodal models evolve.
\end{itemize}

%\fb{Our results suggest that these multimodal models can correctly perform operational and representational tasks on graphs and trees across different structural and aesthetic forms to some degree. GPT-4o, a top performer, achieves a \textbf{87.6\%} accuracy on tree problems but only reaches \textbf{44.7\%} on graphs. This paper contributes another critical capability to the evolving list of qualifications for LLMs and LMMs, while also exposing the difficulties of certain concepts like directed and undirected graph variants for models. However, it is likely these LMMs will only continue to progress. Considering that graphs and trees are fundamental computing concepts, the performance of these models on such problems continues to challenge the circumvention of LLM use through visual representations. Revisiting assessment practices may continue to be a priority for educators and practitioners.}
%\begin{figure}
%    \centering
%    \includegraphics[width=\linewidth]{figs/gpt4fig1.png}
%    \caption{A screenshot of GPT-4V getting the correct answer for an image of a Parsons problem (EvoParsons Haynes-Magyar2022figure4). The response breaks down the problem solving process and even identified the distractors.}  
%    \label{fig:gpt-answer}
%\end{figure}}
%At pass@1, GPT-4V achieves \textbf{77.8\%} accuracy on tree samples and reaches \textbf{28.5\%} on graphs. 
%The robust performance of GPT-4V as a top performer sheds light on the progress of LMMs, which shows acuity on data structures like trees while exposing the complexity of graph theory in directed and undirected variants.

\section{Related Work} 
% In the related work section what we are trying to do is present all the past work that is in accordance with our work and at the same time try to carve out the space where our work fits in ... 1. Presentign past work that aligns with or informs your study, 2. Highlighting the novel aspects of our work

\subsection{Graphs and Trees in Computing Education} 

Graph and tree data structures are core components of most computing education curricula~\cite{joint2013computer, sedgewick2011algorithms}. Considered foundational topics in CS2 data structures and algorithm courses, students are expected to be able to represent and operate various tasks on graphs and trees~\cite{joint2013computer}. Common examples of graph and tree tasks focus on traversals, insertions, deletions, and efficiency calculations on derivatives such as binary trees (BT), binary search trees (BST), directed graphs (DG), and undirected graphs (UG)~\cite{joint2013computer}. Previous work emphasizes the value of teaching graph theory concepts to students of all ages~\cite{gibson2012teaching}; these concepts are often important across multiple domains and industries~\cite{van2010graph, riaz2011applications}.

While critical to computer science, graph theory can be complex to grasp~\cite{dagdilelis1998didagraph}. Prior research has identified popular student misconceptions and errors, such as assuming binary search trees are balanced by default or performing insertions at the wrong node~\cite{karpierz2014mis, zingaro2018identifying, zingaro2018identifying}. Students also struggle with traversing BSTs~\cite{murphy2015bug}, understanding the distinction between heaps and trees,~\cite{danielsiek2012detecting}, and implementing error-free recursive code for operations~\cite{grissom2016paper}. Students especially need help with algorithmic implementation~\cite{medova2019undergraduate}.

Given that graphs and tree problems are notoriously challenging for students, pedagogical strategies and visualization techniques have been devised to support the representation of these concepts. For example, students' understanding can be improved with visualizations, drawing attention to the relationships between nodes, branches, and structure~\cite{Seidametova2021SomeWO, Budiman2018Mobile, sevcikova2016Multimedia, perlin2018chalktalk, byrne1999evaluating}. %Other approaches have aimed to enhance student motivation and mindset to improve retention of these concepts~\cite{chang2017Exploring, Lee2015DataStruct}. %{\color{red}Similarly, when assessing student knowledge, graph and tree problems continue to be represented through a variety of visual formats~\cite{}.} 

Based on this difficulty and visual complexity, these problems are an effective benchmark for testing the spatial reasoning capabilities of mulitmodal models in computing education. Prior work demonstrated that these models can solve text-based Parsons problems based only on images, though those tasks required limited spatial reasoning ability~\cite{hou2024more}. However, graphs can be represented in multiple styles with nodes and links placed in varied ways.

\subsection{Generative AI Threats to Assessment} 

The emergence of LLMs and LMMs has sparked widespread concern among educators regarding effective assessment methods in computing education. Educators are particularly worried about academic integrity and the potential for students to overly rely on generative AI tools, potentially compromising genuine learning~\cite{zastudil2023generative, lau2023from, becker2023programming, prather2023robots, macneil2024imagining}. These concerns have only been amplified by recent research on the wide-ranging capabilities of these models in computing contexts such as solving programming problems~\cite{puryear2022github, wermelinger2023using, finnie-ansley2023my, chen2021evaluating, reeves2023evaluating, hou2024more} and performing as well as students on multiple choice quizzes~\cite{ savelka2023large, savelka2023thrilled}.

While some instructors initially leaned towards prohibiting AI models in the classroom, many believe that 'resistance is futile' and that regardless of sanctions, students would use them~\cite{lau2023from, zastudil2023generative}. Most computing students report being aware of these models and seeking help from them on a regular basis~\cite{hou2023effects, budhiraja2024jarvis}. In response, some practitioners are adopting more structured uses of AI tools in classrooms, allowing students guided access to these models~\cite{Kazemitabaar2023studying, liffiton2023codehelp}, while others are developing problems to circumvent student usage~\cite{denny2023promptly}. Discourse on circumventing LLM usage also includes possibilities of returning to proctored or oral exams~\cite{zastudil2023generative, Joshi2023ChatGPT, susnjak2022chatgpt}, despite known challenges and biases~\cite{gonzalez2020implementation, coghlan2021good, gudino2021remote, cahapay2021problems}. Despite a potential 'arms race' between AI capabilities and instructor interventions, and capabilities continue to improve. More recently, multimodal models have emerged which can process text and image data. A recent paper~\cite{hou2024more} showed that multimodal models can even solve Parsons problems based only on an image of the problem.  % accelerating, whether these models can realistically solve complex vision tasks in fundamental data structures remains unclear. Prior work primarily focuses on LLMs and text-based programming problems. More research is needed on their capabilities to understand how these models, especially multimodal models, threaten assessment.

%TODO add parsons work, preferences?, use to help with 'busy work?', then the solutions are emerging + tovi grossman, some final motivation push...

%\begin{itemize}
%    \item LLMs are great at explanations~\cite{leinonen2023comparing, macneil2023experiences} 
%    \item Concerns about academic dishonesty~\cite{prather2023robots, zastudil2023generative, macneil2023implications, lau2023from, becker2023programming, macneil2023imaginging}. 
%    \item Examples (Parsons~\cite{reeves2023evaluating}, MCQs~\cite{savelka2023thrilled}, Visual Parsons~\cite{hou2023parsons}, etc) 
%    \item Solutions are emerging (Prompt Problems~\cite{denny2023promptly, prather2024interactions} and CodeHelp~\cite{liffiton2023} and work from Tovi Grossman's lab 
%    \item These solutions require substantial work or lock students into specific learning environments. 
%\end{itemize}

\section{Methodology}
% Our Methodology was for why? 
% We needed to justify the selection of Problems 
% There was no previous DataSet
% No followable methodology or structure to test Visual Programming problems

Building on previous research that benchmarks the performance of LLMs and LMMs in computing education~\cite{hou2024more, cassano2023multipl, reeves2023evaluating}, we evaluated the capabilities of LMMs, such as GPT-4V, in solving visual programming problems.

%\fb{This study builds on previous explorations of LMMs such as GPT-4. %The selected models can process text and image data as input with text data as output~\cite{hou2024more}. 
%In that prior work, Hou et al.~\cite{hou2024more} highlighted GPT-4V and Bard's unexplored impact within computing education. We find capable performance in vision tasks on Parsons programming problems. This previous work supports our motivation to generalize LMM performance in multiple domains of computing education.

Recognizing that image-based problems can help mitigate cheating in computing education contexts~\cite{denny2023promptly}, and noting the limited research on evaluating visual problem-solving skills~\cite{hou2024more}, we investigate the ability of LMMs to solve tree and graph data structures problems. We hypothesize that these problems will present unique challenges due to their reliance on spatial reasoning. Our goal is to uncover insights into the interpretative and spatial reasoning abilities of LMMs, representing a significant advancement in understanding their applications in educational contexts. This section describes the creation of our dataset and our evaluation process. %, the specifications for image inputs, and the effective prompting methods utilized in our model evaluation process.

%By focusing on the intricate dynamics of fundamental graph theory, we aim to uncover insights into these multimodal systems' interpretative and reasoning abilities, marking a significant step forward in understanding their applications in educational contexts. We further explain the development of our dataset architecture, image specifications, and effective prompting methods to form the model evaluation process.

%This study aims to investigate the recently deployed large multimodal models (LMMs) that generate natural language output given a text prompt and image. These models, such as Gemini and GPT-4V, have not yet been studied in the context of computing education, but they could have significant impacts. Over a year ago, researchers raised the clarion call that `robots are coming'~\cite{Finnie-Ansley2022Robots}, but in this work we focus on the ability for these new models to interpret and solve visual programming problems. We use Parsons problems as a step toward understanding their capabilities and implications more broadly. To conduct this investigation, we needed to develop a dataset of visual Parsons problems and design a prompt that performs well at solving these problems. In this section, we describe how we developed our dataset, engineered an effective prompt, and formed the process for evaluating the model's performance.% on our dataset. 

\subsection{Creating the Benchmark Dataset}

Data leakage presents significant challenges when evaluating the performance of LLMs and LMMs. These models are typically trained on extensive datasets sourced from the internet, which raises the risk that they may encounter similar problems during both training and testing phases. If the test dataset includes problems that already exist online, the model could leverage its prior knowledge to solve these problems, rather than demonstrating its true problem-solving capabilities. This situation can result in inflated performance metrics, ultimately undermining the validity of the evaluation.

To mitigate this issue, we developed a benchmark dataset comprising graph and tree problems that do not exist on the internet. We created a Python script to programmatically generate these graphs and trees, ensuring diverse yet systematic variations. By guaranteeing that the problems are unique and not previously encountered by the model, we significantly reduce the risk of data leakage and enhance the reliability of our assessments.

In our commitment to advancing the research community, we have made both the generator and the benchmark dataset publicly available on GitHub under an MIT license\footnote{https://github.com/gutbash/lmm-graph-vision}. By providing these resources, we aim to foster collaboration, facilitate replication, and encourage the extension of our findings, thereby establishing a solid foundation for future investigations into the capabilities of large multimodal models in tackling complex programming challenges.

%Given that a dataset of strict graph and tree vision-language tasks does not exist, we form a framework for future multimodal evaluations to curate a dataset of vision-language tasks. We ensure the dataset is modular and allows customization for generalized multimodal evaluation. Finally, we open-source the final framework on GitHub under an MIT license\footnote{Anonymized for Submission}.
% https://github.com/gutbash/lmm-graph-vision

\begin{figure*}
    \centering
    \includegraphics[width=1\linewidth]{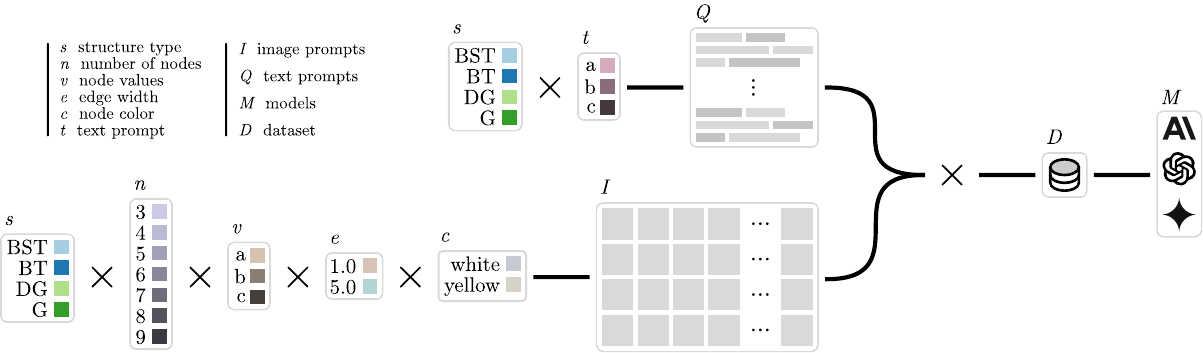}
    \caption{Overview of our benchmark dataset creation process, illustrating the transition from dataset construction to model evaluation. The image prompt set $I$ and text prompt set $Q$ are generated as follows: each image $I_i$ has a combination of attributes where $s$ denotes data structure type, $n$ denotes the number of nodes, $v$ denotes variation of node values, $e$ denotes edge width, and $c$ denotes node color. A text prompt $Q_i$ combines attributes, where $s$ denotes data structure type and $t$ denotes the operational task category. The sets of image and text prompts combine to form the dataset $D$, where they are fed to the model $M_i$.}
    \label{fig:flow}
\end{figure*}

%Given that a dataset of visual Parsons problems does not currently exist, we formed our own dataset guided by recent related research. Based on a recent working group by Ericson et al.~\cite{ericson2022parsons}, we chose four commonly referenced Parsons problem generators, which each varied in visual style. We then drew six Parsons problems from a recent study that assessed LLM performance on text-based Parsons problems~\cite{reeves2023evaluating}. The selected Parsons problems are representative of introductory programming concepts and have all appeared in previous literature. 

\subsubsection{Task Criteria} 
\fb{
We establish a rigorous process for obtaining graph and tree data structures vision tasks. We seek to create a standard, repeatable method to ensure consistent and reliable task selection through clear inclusion and exclusion factors.
 
\begin{enumerate}
\item \textbf{Curricular Grounding}: Subject areas were sourced from the ACM computer science curriculum guidelines~\cite{cs2013} to minimize bias and ensure the selection of relevant topics. %This guides the final subject areas and core topics to focus on.
\item \textbf{Alignment with Core Topics and Learning Objectives}: Evaluation tasks are selected based on their relevance to the core topics and learning objectives outlined in the curriculum guidelines, ensuring that the tasks are contextually appropriate and aligned with educational goals.
%\item \textbf{Alignment with Learning Outcomes}: Tasks are selected based on their alignment with learning outcomes identified in curriculum guidelines to ensure contextual relevance.
\item \textbf{Adherence to Task Standards}: Each task requires multimodal reasoning, including object reasoning, spatial relations, and counting. We designed tasks to (a) maximize open-endedness, (b) minimize unnecessary image prompt details, and (c) reduce dependence on explicit output formats, aligning with best practices for vision-language tasks~\cite{agrawal2016vqa}.%foundational vision-language task research~\cite{agrawal2016vqa}.
\item \textbf{Image Specifications}: Images are standardized to a \textbf{1:1} aspect ratio and \textbf{512x512} resolution, which represents the minimum resolution recommended by the models' API guides at the time of publication. This standardization allows for a systematic assessment of model performance.% across varying features while emphasizing multimodal capabilities.

\end{enumerate}

This comprehensive approach allows a diverse and robust dataset that enables in-depth analysis of model capabilities on graph and tree data structures vision-language tasks.
}

\subsubsection{Task Selection}

The selection of samples is guided based on the \textit{Curriculum Guidelines for Undergraduate Degree Programs in Computer Science}. Specifically, we focus on the topics classified under `Core Tier 1' in the `Fundamental Data Structures and Algorithms' course of the `Algorithms and Complexity (AL)' domain. We are particularly concerned with the following tasks: 

\begin{quote}
    \textbf{Binary search trees:} Common operations on binary search trees such as select min, max, insert, delete, iterate over tree.

    \textbf{Graphs and graph algorithms:} Representations of graphs (e.g., adjacency list, adjacency matrix) and traversals (e.g.: breadth- and depth-first).
\end{quote}

%These tasks are deemed essential knowledge for students in a data structures course. 
We also ensure that the tasks are aligned with core curriculum standards and the learning outcomes for data structures courses. We are particularly focused on the eighth learning outcome:

\begin{quote}
    Solve problems using fundamental graph algorithms, including depth-first and breadth-first search. [Usage]
\end{quote}

Based on these curricular topics and learning objectives, we are able to justify our epistemic decisions. We adopted four structures: binary tree, binary search tree, undirected graph, and directed graph. We also focused on key tasks, such as traversals. This selection process ensures educational relevance for our work. 

%Based on the core topics and learning outcome objectives, this selection process allows us to categorize tasks into four main structures: binary tree, binary search tree, undirected graph, and directed graph. By adhering to these objectives, we established a robust framework for task selection. This selection process validates our dataset's educational relevance and ensures its applicability across similar research endeavors as a useful framework for benchmarks. The structured approach to selecting questions based on these core objectives offers a consistent model for future studies.

\subsubsection{Process for Constructing the Benchmark Dataset}

To construct our benchmark dataset, we followed the process outlined in Figure~\ref{fig:flow}. Initially, three authors sourced problems from academic textbooks, ensuring alignment with our previously established criteria. This step enabled us to identify key variations in the representation of graphs and trees, including node values, node colors, and edge widths. Leveraging these variations, we programmatically generated our benchmark dataset.
%The construction of our dataset follows a layered approach designed to ensure broad coverage of problem types while maintaining high-quality standardization. Three authors initiated the process by sourcing problems from academic textbooks that align with our selection criteria based on core objectives, then refine the problem into a text prompt format based on our task criteria.
%Subsequently, two additional authors categorized each problem by following the criteria and procedure from the previous section.
%We wrote a Python script which used a graph networking libraries such as \textit{networkx} and \textit{matplotlib} to systematically generate a relevant image for a given text prompt to form a complete vision task. This dual authorship filtering and categorization ensures a high fidelity between the dataset content and its educational purpose.
We created a Python script that used graph libraries (i.e. \textit{networkx} and \textit{matplotlib}) to systematically create relevant images for the vision-language tasks. %This multi-authorship approach ensures high fidelity between the dataset content and its educational objectives.

\begin{figure*}
    \centering
    \includegraphics[width=1\linewidth]{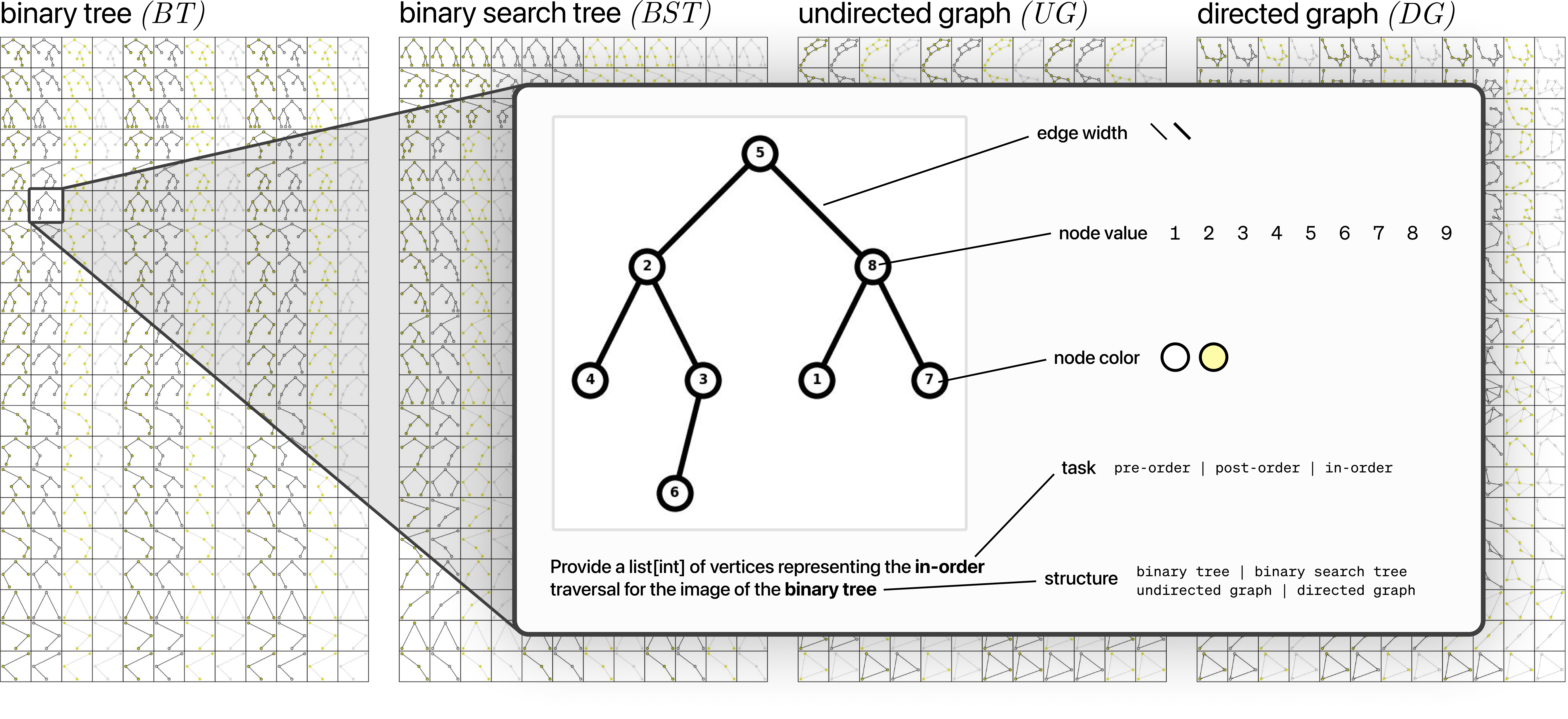}
    \caption{The complete set of images from the dataset.}
    \label{fig:dataset}
\end{figure*}

As shown in Figure \ref{fig:flow}, the resulting structured dataset, denoted as \(D\), consists of vision-language task instances designed to assess the capabilities of multimodal systems. Each instance within \(D\) is a tuple $(I_i,Q_i,A^\text{true}_i),$
%\begin{align*}
%(I_i,Q_i,A^\text{true}_i), \\
%\end{align*}
where \(I_i\) is an image prompt from a set of images $I$ (see Figure \ref{fig:dataset}) depicting a graph or tree, \(Q_i\) is a text prompt that asserts an imperative task relevant to \(I_i\) from a set of text prompts $Q$, and \(A_i^\text{true}\) is the accurate solution or answer to \(Q_i\) with respect to \(I_i\) from a set of ground truth answers $A^\text{true}$. %This construction allows for a targeted evaluation of the models' understanding and reasoning over graph theory vision tasks and their properties.

The graph-related tasks form the subset \(D_G\), composed of individual instances \((I_{gi}, Q_{gi}, A^\text{true}_{gi})\), where each \(I_{gi}\) is an image prompt of a graph structure, \(Q_{gi}\) the accompanying text prompt, and \(A^\text{true}_{gi}\) the ground truth answer. This subset is further divided into tasks concerning undirected graphs \(D_{UG}\) and directed graphs \(D_{DG}\), facilitating an in-depth examination of model performance on each graph type. Tree-related tasks are similarly compiled into subset \(D_T\), with each instance \((I_{ti}, Q_{ti}, A^\text{true}_{ti})\) representing a tree structure question and its answer. This subset is bifurcated into binary tree tasks \(D_{BT}\) and binary search tree tasks \(D_{BST}\), allowing the assessment to account for the nuanced differences in processing and analyzing these particular tree structures.

%The dataset \(D\) holds instances of tasks, \(D_i\), comprised of the tuple \((I_i,Q_i,A_i^\text{true})\). 
Each image $I_i$ has a combination of attributes where $s$ denotes the type of data structure (BST, BT, UG, DG), $n$ denotes the number of nodes (3-9), $v$ denotes variation of node values given a data structure, $e$ denotes rendered edge width (1.0 or 5.0), and $c$ denotes node background fill color (yellow or white). A text prompt $Q_i$ combines attributes, where $s$ denotes data structure type and $t$ denotes the task (i.e. post order, in order, breadth-first search, etc.). %The sets of image and text prompts combine to form the dataset of tasks $D$ where \(|D| = 9072\). These tasks are fed to a given model $M_i$. 
%These attributes lend a hand to the feature diversity of the dataset through our framework and can be scaled for larger evaluations. The level of variety in tasks ensures that the dataset can challenge and assess a model's capability to interpret and solve a wide array of vision tasks.

The dataset is divided equally between graph ($|D_G|=4536$) and tree ($|D_G|=4536$) problems, with a further subdivision into tasks for binary tree, binary search tree, directed graph, and undirected graph categories where $|D_{BT}|=2268$, $|D_{BST}|=2268$, $|D_{UG}|=2268$, and $|D_{DG}|=2268$. The dataset comprises a set of 9072 tasks where $|D|=9072$. These tasks are fed to a given model $M_i$.

Using our framework and carefully selecting features are deliberate choices that enhance the dataset's utility for future research. We aim to create a resource that serves current educational needs and sets a precedent for creating comprehensive vision-language task datasets that are conducive to advancements in the field.

%Recently large multimodal models have been developed~\cite{wang2023review, liu2023hidden, gong2023multimodal, liu2023visual}. While new models are being released on a nearly monthly basis, we selected two highly popular models---OpenAI's GPT-4V and Google's Bard. At the time of writing this paper, GPT-4V is not yet accessible through the API. To ensure a fair comparison, we used the user interface to interact with GPT-4V and Bard. 

\subsection{Prompt Engineering}
\fb{We establish a standard prompt method across all tasks as a zero-shot imperative task to the model in reference to the paired image. The schema consists of a user message with the text and base64 image prompts, followed by the model response as a prediction. We do not include a system message as the Gemini models currently do not support it. %Each prompt is based on a particular graph theory operational or representational task for the model. 
These base prompts are templates that contain two keys to insert information related to the given image prompt. One key is for the given data structure, such as a binary tree, and the other is a vertex node for graph prompts to instruct a model where to start a search traversal.

These prompts align with our task criteria, where no step-by-step instructions are given to perform the task correctly, remaining open-ended. The prompts also do not indicate the data structure's content other than the identification of the structure's type, minimizing image details. The only exceptions to this are depth-first search and breadth-first search tasks where a starting vertex was provided. 
The prompt does not explicitly mention the programming language, but the type annotation is in Python. We specify such a format to allow the prediction to be extracted using regular expressions.% to filter from the supplementary response content. 
Three base text prompts take this described form:

\par\smallskip\smallskip
\begin{custombox}
{Provide a list[int] of vertices representing the \textit{\{pre-order, in-order, post-order traversal\}} for the image of the \textbf{\{structure\}}.}
\end{custombox} \vspace{-2pt}
%\begin{custombox}
%{Provide a list[int] of vertices representing the pre-order traversal for the image of the \textbf{\{structure\}}.}
%\end{custombox}
%\begin{custombox}
%{Provide a list[int] of vertices representing the in-order traversal for the image of the \textbf{\{structure\}}.}
%\end{custombox}
\begin{custombox}
{Provide a dict[int, list[int]] of vertices and their edges representing the adjacency list for the image of the \textbf{\{structure\}}.}
\end{custombox}\vspace{-2pt}
\begin{custombox}
{Provide a list[int] of vertices representing the \textit{\{depth-first, breadth-first\}}  search traversal for the image of the \textbf{\{structure\}} starting from vertex \textbf{\{vertex\}}.}
\end{custombox}
%\begin{custombox}
%{Provide a list[int] of vertices representing the breadth-first search traversal for the image of the \textbf{\{structure\}} starting from vertex \textbf{\{vertex\}}.}
%\end{custombox}

Preemptive sample evaluations on Gemini 1.0 Pro Vision show minimal differences in accuracy compared to the six prompts tested, so our premise around how to prompt the model focuses on zero-shot learning. This involves evaluating model performance without examples. This simpler prompting technique is probably more likely among computing students than more advanced techniques. Instructions expect an output format in Python type annotations, though they do not express an explicit language intent. The prompt specifies the existence of an image and its general subject but does not expose any more information unless necessary for the task. These specifications align with our established standards and ensure isolated consistency in predictions across different categories.
}

\subsection{Model Selection}
\fb{To select models for this study, we aimed to align with established research practices while incorporating diversity and leveraging the most state-of-the-art models. Consequently, we first used OpenAI's GPT-4V model due to its widespread use in LMM benchmarks and popularity among both researchers and consumers (via ChatGPT). Initially, we conducted preliminary tests comparing GPT-4V with Google’s Bard (prior to its rebranding as Gemini) to broaden our investigation. During these tests, we observed that Bard demonstrated inferior performance in our evaluations. However, with the subsequent release of Google’s Gemini Pro 1.0 and 1.5 models---which officially replaced Bard---we observed performance comparable to GPT-4V in our pilot studies, prompting us to include both GPT-4V and Gemini as primary systems in our research. Just before the first analysis, the Claude 3 model family was released with impressive benchmarks which prompted us to include it. After the first analysis, GPT-4o was also released and we repeated the analysis for that model. Finally, all parameter values were left to default since not all models shared them, except for temperature. We elected a temperature of 1.0 to be used for the set of models since this was the most common default temperature. %Finally, the release of the Claude 3 model family and impressive benchmarks prompted us to include it in our evaluation. 
This set of models aims to constitute a good representation of current state-of-the-art research and consumer multimodal systems.}

A given multimodal model \(M_i\) from a set of models \(M\) is a function to process a given vision-language task, mapping the input pair of an image and its text prompt to an invoked prediction. Formally, for a task \((I_i,Q_i,A_i^\text{true})\), the model function is applied as \(A^\text{model}_i = M_i(I_i, Q_i)\) where \(A^\text{model}_i\) is the model's predicted answer. The objective of \(M_i\) is to return an accurate prediction for \((I_i, Q_i)\). See Appendix \ref{appendix:performance-details} for a further breakdown of performance.

\subsection{Evaluation} % Gemini (June), 4o in June, March (GPT-4V) 

The first evaluation took place on April 17, 2024 and included the Claude family of models and GPT-4V. As other models---such as GPT-4o, Gemini 1.5 Flash, and Gemini 1.5 Pro---became available, a second evaluation was completed for those models on June 13th, 2024. 
%The first evaluation was conducted over a two-week period ending near late March 2024. Gemini 1.5 Flash, Gemini 1.5 Pro, and GPT-4o were evaluated in early June 2024, while and Claude Sonnet model was evaluated in X. %When GPT-4o was released, we conducted an evaluation for that model, which occurred in early June 2024. 
OpenAI models had few rate limitations, and the evaluation was completed in around 45 minutes. Gemini 1.5 Pro was the earliest released model in a preview state and, therefore, was slower to complete, at around three days. Gemini 1.0 Pro Vision was completed over two hours. Finally, the Claude 3 family of models completed each task quickly, but due to token rate limitations, it ended up taking around the same time as Gemini 1.0 Pro Vision at around two hours each. Results were saved to CSV files, where cleaning and parsing were done to ensure accurate measurements.

To assess the accuracy of the model responses, we define a performance metric, \(P\), which quantitatively evaluates the correspondence between \(A^\text{model}_i\) and \(A^\text{true}_i\). The metric \(P\) is a function \(P(A^\text{true}_i, A^\text{model}_i)\) that outputs a binary true or false value representing the correctness of the model-generated answer to the ground truth. In practice, our program uses regular expressions to extract the model's answer to compare with the ground truth. %{\color{red}See \ref{sec:performance} for details on how we obtain performance over sets of tasks.} 
The performance metric for each model is measured in both pass@1 and pass@3 accuracy, which shows if the model answered correctly on the first attempt or within three attempts respectively.

Our analysis requires a systematic comparison across models and task categories. The comparative function is thus articulated to discern the model that exhibits superior performance on the designated tasks. For each category of graph \(G\) and tree \(T\) tasks, we compute the mean performance across all relevant instances, delineating specific measures for undirected graphs \(UG\), directed graphs \(DG\), binary trees \(BT\), and binary search trees \(BST\). These measures facilitate a nuanced comparison across models, revealing their strengths and weaknesses. Thus, let \(M = \{M_1, M_2, \dots, M_n\}\) be the set of models we compare. We form separate performance matrices for each category and subcategory with graph tasks being \( P^{model}_G(type)\), where \(type\) can be UG or DG, and with tree tasks being \(P^{model}_T(type)\) where \(type\) can be BT or BST. Finally, function \(C\) takes the performance matrices and compares them to find the best-performing model for each set of tasks:

\begin{equation}
\begin{aligned}
C(M, type) = \text{arg max}_{M_i \in M} P^{M_i}_{type} \\
\end{aligned}
\end{equation}

{
\begin{table*}[!ht]
\centering
\label{table:combined_model_accuracy_structure}
\caption{Zero-shot pass@$k$ accuracy (\%) by model and data structure: $BT$ denotes binary tree, $BST$ denotes binary search tree, and $T$ denotes their combined set. $UG$ denotes undirected graph, $DG$ denotes directed graph, and $G$ denotes their combined set. Bolded values indicate highest accuracies.}
\small
\begin{tabularx}{0.85\linewidth}{@{ }l*{13}{S}@{ }}
\toprule
& \multicolumn{2}{c}{$BT$} & \multicolumn{2}{c}{$BST$} & \multicolumn{2}{c}{$T$ (Overall)} && \multicolumn{2}{c}{$UG$} & \multicolumn{2}{c}{$DG$} & \multicolumn{2}{c}{$G$ (Overall)} \\
\cmidrule(lr){2-3} \cmidrule(lr){4-5} \cmidrule(lr){6-7} \cmidrule(lr){9-10} \cmidrule(lr){11-12} \cmidrule(lr){13-14}
{Model / pass@$k$} & {$k = 1$} & {$k = 3$} & {$k = 1$} & {$k = 3$} & {$k = 1$} & {$k = 3$} && {$k = 1$} & {$k = 3$} & {$k = 1$} & {$k = 3$} & {$k = 1$} & {$k = 3$} \\
\midrule
GPT-4o & {\textbf{73.3}} & {\textbf{82.7}} & {82.3} & {\textbf{92.5}} & \textbf{77.8} & \textbf{87.6} && {38.5} & {49.7} & {27.5} & {39.6} & {33.0} & {44.7} \\
GPT-4 Vis. Prev. & {71.8} & {78.4} & {\textbf{83.7}} & {90.6} & {77.8} & {84.5} && {38.0} & {50.9} & {19.0} & {29.0} & {28.5} & {40.0} \\
Gemini 1.5 Pro & {65.5} & {71.3} & {67.5} & {70.8} & {66.5} & {71.1} && {\textbf{52.2}} & {\textbf{56.3}} & {46.2} & {51.2} & {49.2} & {53.8} \\
Gemini 1.5 Flash & {64.0} & {65.3} & {72.0} & {75.3} & {68.0} & {70.3} && {50.7} & {55.3} & {\textbf{53.3}} & {\textbf{57.1}} & \textbf{{52.0}} & \textbf{{56.2}} \\
Gemini Pro Vision & {56.5} & {59.5} & {57.0} & {59.5} & {56.8} & {59.5} && {32.1} & {34.5} & {34.3} & {35.8} & {33.2} & {35.2} \\
Claude 3 Opus & {27.0} & {35.7} & {50.4} & {63.1} & {38.7} & {49.4} && {16.3} & {22.6} & {20.1} & {28.8} & {18.2} & {25.7} \\
Claude 3 Sonnet & {31.6} & {41.8} & {41.1} & {53.4} & {36.4} & {47.6} && {25.4} & {31.9} & {27.6} & {35.3} & {26.5} & {33.6} \\
Claude 3 Haiku & {23.7} & {29.8} & {58.2} & {64.6} & {41.0} & {47.2} && {19.8} & {23.3} & {23.3} & {29.0} & {21.6} & {26.2} \\
\bottomrule
\end{tabularx}
\end{table*}
}

\begin{figure*}
    \centering
    \includegraphics[width=1\textwidth]{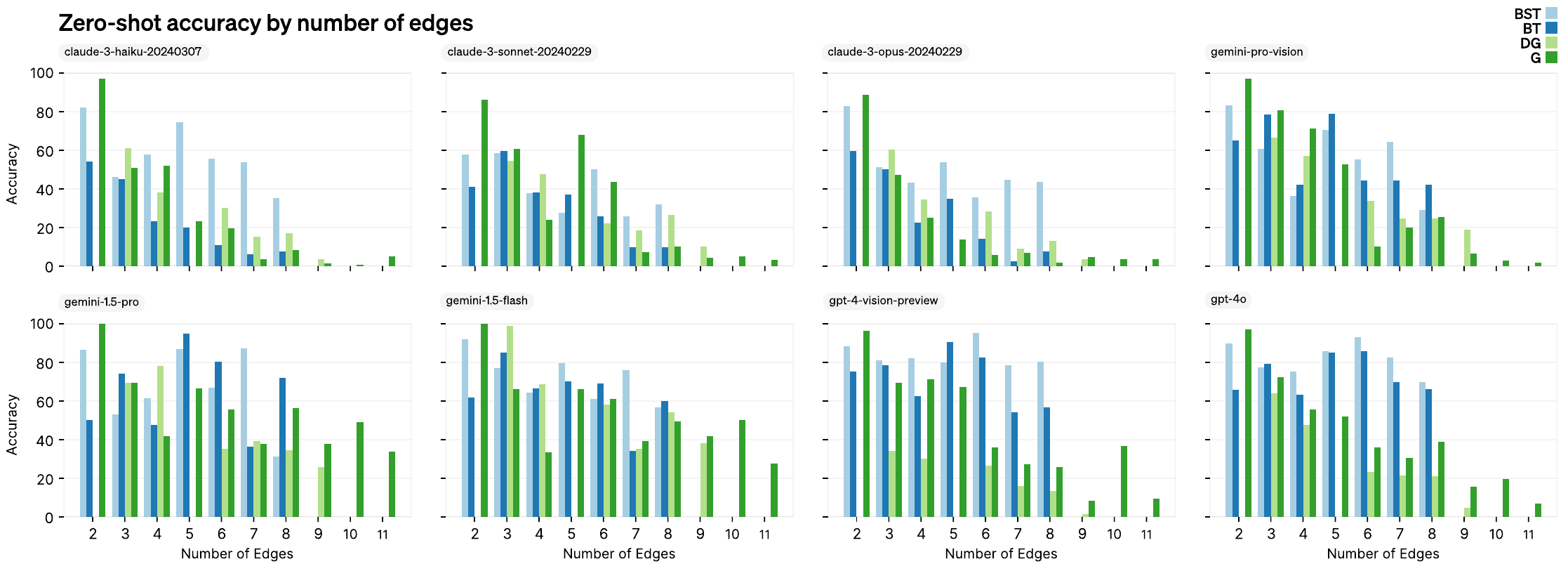}
    \caption{Zero-shot accuracy overall (\%) by model, number of edges, and structure. Consistent with the feature importance analysis, accuracy decreases as the number of edges increases. }
    \label{fig:edge-numbers}
\end{figure*}

\subsubsection{Feature Importance}

To address RQ2 and RQ3, we trained a classification model on evaluation results to examine the influence of specific features on model performance. The feature set included experimentally varied aesthetic aspects (e.g. edge width, node colors, and node value variations) and structural features (e.g. number of nodes, node values, graph density, etc.) to capture the visual diversity of graph and tree representations. Additionally, we used Deep Graph Learning (DGL) to compute augmented features, such as graph density and degree, which may impact accuracy in structurally complex tasks. Finally, the task type, determined by the six prompt categories, was included as a feature to understand how different tasks affect multimodal model performance.

To minimize overfitting and enhance model interpretability, we applied Principal Component Analysis (PCA) to image features captured with EfficientViT (efficientvit\_l3.r384\_in1k), reducing noise and eliminating less relevant features. Finally, we fit a logistic regression model to the refined feature set for each multimodal model, achieving F1 scores ranging from 0.693 to 0.843, with an average score of 0.791. This high performance suggests that the features are both important and predictive. See Appendix \ref{appendix:classification} for details on classification model performances.

%The data structure type was an additional feature. Finally, we included the task as a feature which was extracted from the six prompts. We used PCA to do feature selection and eliminate irrelevant features from the model to avoid overfitting. Next, we fit a logistic regression model to the selected features for each multimodal model. The F1 scores ranged from 0.693 to 0.843 with an average score of 0.791.  

%In addition to descriptive analysis, we trained a classification model on the evaluation results to understand the feature importance. We used the features edge width, node colors, structure variants, node value variations. 

%In addition to descriptive statistical analysis, we train a classification model on the evaluation results. We find feature importance from graph features using Deep Graph Learning (DGL) to extract structural features, principle component analysis (PCA) for image features through EfficientViT (efficientvit\_l3.r384\_in1k), edge width, node colors, structure variants, node value variations, and text prompts through TF-IDF vectorization followed by SVD for dimensionality reduction. %See Appendix \ref{sec:features} for details of the feature engineering and Appendix \ref{sec:classification} for training results.

\section{Results}

%From the results, we form explanations for each research question of overall performance, followed by the influence of structural and aesthetic features. Our analysis reveals insights into model accuracy on the dataset and further illustrates the importance of distinct features.

We measure accuracy for undirected and directed graphs in addition to binary and binary search trees via pass@$k$ performance. Our findings indicate GPT-4o performs at a higher overall accuracy than any other model in the selection for trees, followed by Gemini 1.5 Flash for graphs. See Appendix \ref{appendix:token-analysis} for a detailed response analysis.
%However, an in-depth look shows that GPT-4V is surpassed in accuracy by almost every other model regarding directed graphs. In addition, Performance between the Claude 3 models varies unexpectedly.

% Observations:\td{It usually provides a general definition of the task, a step-by-step attempt, and an explanation of reasoning.The Claude 3 model family follows similarly, albeit slightly less comprehensive. However, this changes with the Gemini model family, as any auxiliary element of the response is seldom given except for the sole prediction. This starkly differs from the common response style but still accomplishes comparable accuracy to other models, such as Claude 3 Opus, for far fewer tokens.     An analysis of the GPT-4V response style shows consistency in average token length \td{STD DEVIATION}. }

\begin{figure*}
    \centering
    \includegraphics[width=1\textwidth]{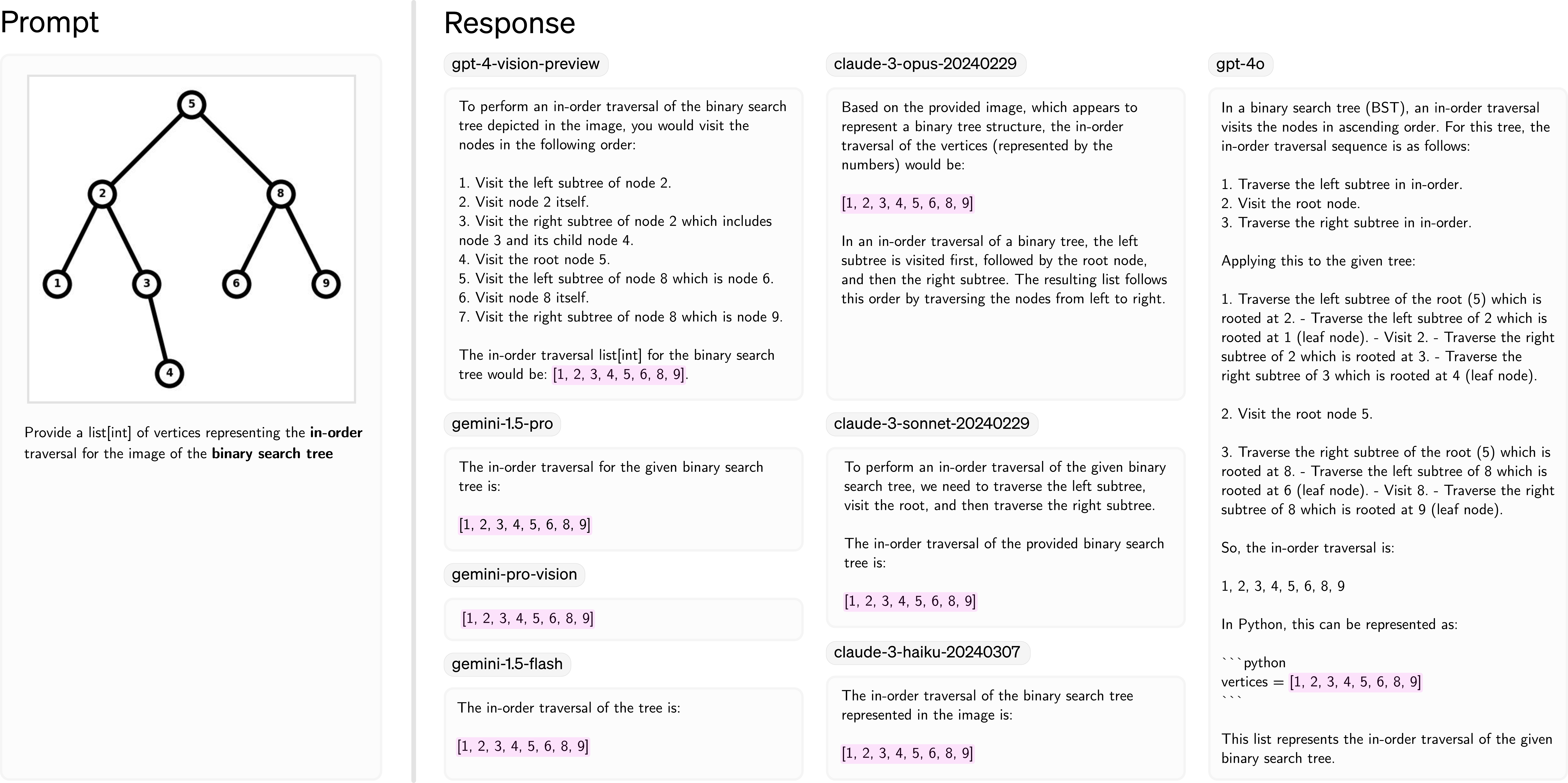}
    \caption{Example model responses to a in-order traversal of a binary search tree. The extracted predictions are highlighted.} %TODO add the date}
    \label{fig:example}
\end{figure*}

\subsection{RQ 1 Overall Performance}

\subsubsection{Tree Performance}
GPT-4o is the top performer on trees with \textbf{77.8\%} pass@1 and \textbf{87.6\%} pass@3 accuracy, which is nearly tied with GPT-4V performance at \textbf{77.8\%} pass@1 and \textbf{84.5\%} pass@3 accuracy. GPT-4o surpasses the average range of performance between models on trees, operating at roughly 20\% higher accuracy than the next best after GPT-4V. Additionally, Gemini 1.5 Pro is only more accurate than Gemini 1.5 Flash by a small margin, with scores of \textbf{71.1\%} (+0.8\%) pass@3. On the lower end, Claude 3 Haiku, the smallest model out of the family, achieves greater pass@1 performance than its siblings at \textbf{41.0\%}, an average of +3.5\% higher than its siblings. However, pass@3 accuracy is surpassed by Claude 3 Opus with \textbf{49.4\%}, an average of +2.0\% higher than its siblings. %The standard deviations for accuracy between all models on trees are ±14.4 pass@1 and ±13.0 pass@3, showing a range of performance across models.

\subsubsection{Graph Performance}
The top performer on the set of graphs is split between Gemini 1.5 Flash, which achieves a top pass@3 accuracy of \textbf{56.2\%}, and Gemini 1.5 Pro, which reaches a top pass@3 accuracy of \textbf{53.8\%}. %Like tree performance, Gemini 1.5 Pro attains higher accuracy than Gemini 1.0 by a small margin, with \textbf{33.7\%} (+0.5\%) pass@1 and \textbf{35.6\%} (+0.4\%) pass@3. 
GPT-4o is the next best performer, reaching a top pass@3 accuracy of \textbf{44.7\%}, slightly ahead of GPT-4V's pass@3 accuracy of \textbf{40.0\%}. For the Claude 3 model family, Sonnet obtains the highest performance across pass@k of \textbf{26.5\%} pass@1, an average of +6.6\% higher than its siblings, and \textbf{33.6\%} pass@3, an average of +7.7\% higher than its siblings. %Unlike the set of trees, graph performance contrasts in the accuracy range between models with standard deviations of ±5.7 pass@1 and ±5.1 pass@3.

\subsection{RQ 2 Structural Features}
%From each model's top 30 important features in Figure \ref{fig:features}, we analyze the influence of structural features representing structural data. 

%Feature importance affords insight into the impact of a given feature towards a target value, in this case, \(P(A^\text{true}_i, A^\text{model}_i)\) where \(A^\text{true}_i = A^\text{model}_i\). 

The structural features were often the most important features. See Appendix \ref{appendix:feature-engineering} for feature engineering details.

\subsubsection{Number of Edges}
The number of edges was the most important feature in terms of model performance. Further analysis of the coefficients suggests a positive influence at a fewer edges and approaches a negative influence as the number of edges increases.

\subsubsection{Degree Histograms}
The distribution of \textit{in-degree vertices} across all vertices in the graph had a strong influence with a slight positive influence at lower distributions and negative influence at higher distributions. %at different distributions at varying coefficients but slightly follow the trend of positive influence at lower distributions and negative influence at higher distributions.}
Similarly, the distribution of \textit{out-degree vertices} exhibit less influence than in-degree features, but they generally adhere to a similar trend as the in-degree features.

\subsubsection{Number of Nodes and Layout}
The number of nodes and layout of a graph or tree exhibited a trend of positive coefficients at a smaller number of nodes and negative coefficients at a higher number of nodes, which aligns with what is observed in Figure~\ref{fig:edge-numbers}.

%Features present the number of nodes and layout of a graph or tree as 'gen id'. The graph or tree displays mostly consistent trends of positive coefficients at a smaller number of nodes and negative coefficients at a higher number of nodes. Details of the findings can be found in Figure~\ref{fig:edge-numbers}.

\subsubsection{Density}
The number of edges determines a graph's density compared to the maximum number of edges possible for a given graph. Our analysis suggests a positive influence at lower and a negative influence at higher densities. %The bins of densities for the dataset suggest a positive influence at lower and a negative influence at higher densities.

\subsection{RQ 3 Aesthetic Features}
%In our analysis of the top 30 important features in Figure \ref{fig:features}, we seldom find the influence of aesthetic features on a given model.

Based on our logistic regression analysis, the aesthetic features were ranked less important than structural features. The \textit{edge width} varied in influence and showed inconsistent impact across width levels. This suggests that models are relatively robust in recognizing the edge connections. 
%\subsubsection{Edge Width}
%the edge width ('edge width n') features vary in influence and show inconsistent impact on different width levels. This suggests the models are robust in their acuity for recognizing different visual representations of node connections.
%\subsubsection{Node Color}
The \textit{node color} is represented as three features that form the RGB value as a color channel. Results show that, similar to edge width, there is a limited influence on models. However, signs from the GPT-4V model describe a negative influence of an increased `node rgb blue channel` value. From the node colors used in the dataset, we surmise that since the white color has a higher value in the blue channel, using white as a node color on a white background may yield worse results than using a contrasting color like yellow on a white background.

\subsection{Additional Observations}

Models performed best on binary search trees, likely due to their consistent structure. Vision transformers, which process images as patches from the top-left to the bottom-right~\cite{touvron2022three}, may help explain this trend. In tree structures, the root vertex consistently appears at the top of the image, aligning with the patch encoding sequence and providing a stable spatial reference. In contrast, graphs feature more arbitrary spatial arrangements, with starting vertices positioned unpredictably across the image. This spatial inconsistency may account for the models' lower performance on graph datasets.

Directed graphs present additional challenges due to their unique structural features, including arrows that models must accurately perceive. Claude 3 Opus outperformed other models on directed graphs, achieving higher accuracy than on undirected graphs. These findings suggest that directed graphs require greater spatial reasoning, making them more challenging for models.

%The binary tree set uses an arbitrary order of node values for each layout which may be less predictable for models. Performance with binary search trees suggests that performance is similar; however, due to a binary search tree's unique parent-child value properties, models may have a slight edge in more predictable rules and possible value order. 

%Directed graphs have unique structural properties, and visually, they have arrows which must be perceived by the models. Claude 3 Opus was the only model with higher accuracy on the directed graph set than the undirected graph set. With these observations, we gather that directed graphs challenge the models' spatial reasoning more than the other representations. 

%Vision transformers typically encode images in patches from the top left to the bottom right~\cite{touvron2022three}. This might account for the difference in performance between the graphs and trees. In trees, initial attention is typically placed on the root vertex, which always starts at the top, a region of space consistent between the images and tokenized patches. Conversely, the starting vertex has more arbitrary regional positions in space on a graph. This spatial inconsistency may lead to more unpredictable model reasoning.

Figure~\ref{fig:example} shows examples of model responses which in some cases may shed light on their reasoning processes. 

%Finally, observe Figure ~\ref{fig:example} for examples of model responses which can indicate their reasoning processes.

\section{Discussion}

Recent research has shown that generative AI tools like ChatGPT can tackle a broad range of programming tasks, especially at introductory levels (CS1)~\cite{cipriano2023gpt3, savelka2023thrilled, ouh2023chatgpt, savelka2023large, hou2024more, wermelinger2023using, koutcheme2023automated, fan2023automated}. These trends have important implications for pedagogy and assessment. Our study extends this body of work in two key ways: by examining advanced data structures problems and by evaluating the spatial reasoning capabilities of multimodal models (LMMs).

Our findings show that multimodal models consistently solve tree problems across a range of visual presentations, including variations in structure and aesthetics, achieving performance levels sufficient to pass traditional assessments. This raises critical concerns about students' ability to leverage these tools as a shortcut, bypassing the development of conceptual understanding. While graph problems showed lower accuracy rates, the use of a simple zero-shot prompting approach suggests a lower bound. Savy students could refine the prompts or use more advanced prompting techniques. Importantly, minimal prompting can still yield strong results, suggesting students don't need extensive prompt-engineering skills to benefit from these tools~\cite{wermelinger2023using, denny2023promptly, prather2024widening, bull2023generative}.

%Our findings demonstrate that multimodal models can consistently solve tree problems across various visual presentations, including structural and aesthetic variations, at a competency level sufficient for passing assessments. This raises concerns that students could leverage these tools to bypass conceptual understanding. While graph problems showed lower accuracy rates, the use of a zero-shot prompting approach suggests a lower bound. Savy students could refine the prompts or use more advanced prompting techniques. Importantly, minimal prompting can still yield strong results, suggesting students don't need extensive prompt-engineering skills to benefit from these tools~\cite{wermelinger2023using, denny2023promptly, prather2024widening, bull2023generative}.

\subsection{Transforming Traditional Assessment}

Our findings challenge the prevailing reliance on traditional assessment methods in computing education, particularly as generative AI continues to evolve. While recent work has suggested that diagram-based and visual problems could counteract students’ over-reliance on text-based AI tools~\cite{cipriano2024picture, denny2023promptly, golesteanu2024chatgptpasstheorycomputing}, our results suggest that this strategy is unsustainable. Multimodal models already demonstrate significant spatial reasoning capabilities, and their ability to process and solve complex visual problems, such as tree structures, suggests that these stopgap measures will soon lose their efficacy. Even video-based problems, which have been proposed as a more robust alternative~\cite{cipriano2024picture}, are unlikely to hold up as  models improve. % advance further. 

In parallel, some educators have turned to AI detection tools in an attempt to identify when students misuse generative AI. These tools rely on statistical methods that compare human and AI writing patterns. However, this approach is flawed. Statistical resemblance alone is not reliable evidence of authorship. Just as writing that mimics the style of a famous author does not constitute plagiarism, content that resembles AI-generated patterns does not necessarily indicate AI authorship. In the context of computing, the prevalence of design patterns, coding conventions, and commonly used libraries only increases the likelihood of false positives. %, further undermining the effectiveness of these tools.

Both of these attempts to counter AI’s influence on assessment---whether through visual-based problems or detection tools---offer only fleeting solutions. Educators are now, perhaps reluctantly, being forced to adopt what some have long advocated for---innovative assessment models like ungrading~\cite{blum2020ungrading} or project-based learning that encourage active, meaningful engagement and prioritize mastery over performance. This shift, once considered an ideal, is now becoming a necessity in the face of rapidly evolving AI tools. By reimagining assessment to emphasize curiosity and growth, we can create a more resilient approach to learning driven by purpose. This is not merely a pedagogical shift, but a moral imperative. Preparing the next generation of computing professionals requires assessments that promote critical engagement with AI systems, enabling students to navigate their limitations and biases.

\subsection{New Opportunities for Pedagogy}

While the focus of this work is on the implications for assessment,  LMM capabilities could also provide benefits for computing students and educators. As CS2 topics like data structures are particularly challenging~\cite{karpierz2014mis, zingaro2018identifying} and often contribute to `weeding out' students who lack prior experience in computer science, disproportionately impacting traditionally underrepresented students~\cite{hatfield2022introductory}. Enhanced AI tools could improve access to complex computing concepts like trees and graphs, potentially offering these students an extra layer of support. As these models advance, they could also facilitate the creation of digital teaching assistants that deliver high-quality scaffolded help~\cite{denny2024desirable}. More specifically, future work could investigate the potential for LMMs to produce code based on a graph or tree. This would provide help to students that struggle with graph and tree algorithms~\cite{medova2019undergraduate}.

\section{Limitations}

\fb{

There are several limitations to our work. First, the tree and graph data structures used do not cover all possible data structures, and the operations evaluated are limited to core topics typically found in fundamental data structures courses. Consequently, these findings may not be generalized to more advanced topics in data structure.
Second, the generated data structures may not encompass the full range of relevant attributes for graphs and trees. For instance, exploring additional edge widths and node colors could provide further insights. Therefore, investigating a broader array of structural and aesthetic features is warranted.
In addition, while advanced prompting methods—such as self-critique, chain-of---thought, prompt chaining, and few-shot learning---can enhance performance in reasoning tasks, our study focused on demonstrating how students might use models with basic zero-shot approaches. This choice reflects the assumption that students may not be familiar with these advanced techniques.
The model parameters also remained consistent across the evaluations, representing the default values that can be found when using these models for generalized purposes. It is almost certain that the performance varies greatly depending on the modification of certain parameters such as temperature and sampling coefficients across a matrix of these possible combinations.
Finally, the models assessed in this study are continually evolving. To facilitate ongoing evaluation of future models, we have open-sourced our benchmarks and generation tools.}

\section{Conclusion}

In this paper, we investigated the capabilities of large multimodal models (LMMs) to solve complex, visually represented data structure problems. Through the development and evaluation of a diverse benchmark dataset of graph and tree structures, we assessed the performance of various LMMs, including the GPT-4 and Gemini models, in tackling spatially and visually challenging tasks. Our results reveal that while certain models, such as GPT-4o and Gemini 1.5 Flash, performed well on tree and graph problems respectively, there remain considerable variations in accuracy that are influenced by structural complexity, such as edge density. These findings underscore the evolving capabilities of LMMs in processing visual and spatially complex problems, which has important implications for computing education. As generative AI continues to advance and reshape computing education, it becomes essential for educators to adapt assessment methods and foster environments that promote intrinsic motivation.


\bibliographystyle{ACM-Reference-Format}
\bibliography{sample}

%\input{appendix.tex}

%\section{Qualitative Examples}
%In this section, we share qualitative examples of model predictions.


\appendix

{\scriptsize
\begin{table*}[ht]
\centering
\caption{Analysis of model response on a character, word, and token (cl100k\_base) basis.}
\label{tab:tokens}
\begin{tabularx}{\textwidth}{@{}l*{3}{p{1.45cm}}*{3}{p{1.45cm}}*{3}{p{1.45cm}}@{}}
\toprule
 & \multicolumn{3}{c}{Characters} & \multicolumn{3}{c}{Words} & \multicolumn{3}{c}{Tokens} \\
\cmidrule(lr){2-4} \cmidrule(lr){5-7} \cmidrule(lr){8-10}
{Model} & {Total} & {Avg ± SD} &  {Min–Max} & {Total} & {Avg ± SD} & {Min–Max} & {Total} & {Avg ± SD} &  {Min–Max} \\
\midrule
GPT-4o & {7 239 477} & {798 ±364} & {32–3101} & {1 258 965} & {139 ±63} & {5–487} & {2 137 607} & {236 ±108} & {14–930} \\
GPT-4 Vis. Prev. & {5 068 618} & {559 ±289} & {32–2054} & {907 905} & {100 ±53} & {6–360} & {1 376 664} & {152 ±79} & {8–561} \\
Gemini 1.5 Pro & {1 077 074} & {119 ±183} & {9–1514} & {210 505} & {23 ±32} & {3–297} & {432 258} & {48 ±54} & {9–486} \\
Gemini 1.5 Flash & {635 233} & {70 ±35} & {9–1321} & {130 953} & {14 ±7} & {3–244} & {297 321} & {33 ±16} & {9–421} \\
Gemini Pro Vision & {266 413} & {29 ±25} & {10–231} & {73 249} & {8 ±6} & {3–46} & {218 407} & {24 ±17} & {9–104} \\
Claude 3 Opus & {3 590 616} & {396 ±214} & {9–1955} & {647 348} & {71 ±40} & {3–369} & {1 038 594} & {114 ±64} & {9–547} \\
Claude 3 Sonnet & {3 957 662} & {436 ±206} & {65–1984} & {700 621} & {77 ±37} & {13–343} & {1 078 054} & {119 ±64} & {21–581} \\
Claude 3 Haiku & {1 311 315} & {145 ±90} & {21–894} & {243 972} & {27 ±16} & {7–164} & {421 265} & {46 ±27} & {19–259} \\
\bottomrule
\end{tabularx}
\end{table*}
}

\section{Evaluation Results}
\label{appendix:token-analysis}
In Table~\ref{tab:tokens}, we share additional analysis of model evaluations. Across the set of all eight models, they constitute over 72,576 responses in total. The response of each model is unique in length, style, and other characteristics.

\section{Performance Details}
\label{appendix:performance-details}
The overall performance for graph tasks, \( \text{Performance}_{M_i}^{G} \), is calculated as follows:

\begin{align*}
\small{\text{Performance}_{M_i}^{G} = \frac{\text{Performance}_{M_i}^{UG} \cdot |D_{UG}| + \text{Performance}_{M_i}^{DG} \cdot |D_{DG}|}{|D_{UG}| + |D_{DG}|}} \\
\end{align*}

where \( \text{Performance}_{M_i}^{UG} \) and \( \text{Performance}_{M_i}^{DG} \) are the respective model performances for undirected graphs and directed graphs defined as:

\begin{align*}
\text{Performance}_{M_i}^{UG} &= \frac{1}{|D_{UG}|} \sum P(A_{ugi}^{true}, M_i(I_{ugi}, Q_{ugi})) \\
\text{Performance}_{M_i}^{DG} &= \frac{1}{|D_{DG}|} \sum P(A_{dgi}^{true}, M_i(I_{dgi}, Q_{dgi})) \\
\end{align*}

Similarly, the overall performance for tree tasks, \( \text{Performance}_{M_i}^{T} \) are calculated as follows:

\begin{align*}
\small{\text{Performance}_{M_i}^{T} = \frac{\text{Performance}_{M_i}^{BT} \cdot |D_{BT}| + \text{Performance}_{M_i}^{BST} \cdot |D_{BST}|}{|D_{BT}| + |D_{BST}|}} \\
\end{align*}

where \( \text{Performance}_{M_i}^{BT} \) and \( \text{Performance}_{M_i}^{BST} \) are the respective model performances for binary trees and binary search trees defined as:

\begin{align*}
\text{Performance}_{M_i}^{BT} &= \frac{1}{|D_{BT}|} \sum P(A_{bti}^{true}, M_i(I_{bti}, Q_{bti})) \\
\text{Performance}_{M_i}^{BST} &= \frac{1}{|D_{BST}|} \sum P(A_{bsti}^{true}, M_i(I_{bsti}, Q_{bsti})) \\
\end{align*}

%\section{Dataset Architecture}
\section{Feature Engineering}
\label{appendix:feature-engineering}
This section details the process and rationale behind the extraction and preparation of the features used to evaluate the performance of the multimodal models. Each feature is derived from various properties of each vision language task sample. The feature engineering process used in the study involves several techniques that transform the data into a suitable format for classification model training. Below is a detailed breakdown of these techniques.

\subsection{Graph Features}
{Graph features are extracted from the adjacency list of each graph, including the number of nodes and edges, graph density, and degree statistics. These features are one-hot encoded and binned to capture structural nuances in the graph data.

\subsubsection{Data Structure Type}
Each structure category, $BT$, $BST$, $G$, and $DG$ is one-hot encoded into a vector.

\subsubsection{Node Value Variation}
The variation identifier representing a unique set of node values for a given graph is one-hot encoded into a vector.

\subsubsection{Generation}
The generation identifier representing a unique spatial graph layout is one-hot encoded into a vector.

\subsubsection{Node Count}
Each graph's node count is converted into a one-hot encoded vector. The length of this vector is determined by the maximum number of nodes across the set of graphs.

\subsubsection{Edge Count}
Similar to node count, the number of edges is also one-hot encoded, where each position in the vector indicates the presence of a specific number of edges, up to the maximum found in the dataset.

\subsubsection{Graph Density}
The density of a graph is calculated as the ratio of the number of edges to the number of possible edges in the graph. Density is calculated differently for directed and undirected graphs. For directed graphs, the formula used is

\begin{align*}
    \frac{edges}{edges \cdot (nodes-1)} \\
\end{align*}

and for undirected graphs it is

\begin{align*}
    \frac{2 \cdot edges}{edges \cdot (nodes-1)}. \\
\end{align*}

The resulting density measure is binned into predefined categories and one-hot encoded.

\subsubsection{Average Node Degree}
The average degree of nodes in the graph represents the average number of connections each node has. This is calculated by dividing the total number of edges by the number of nodes. The result is binned into predefined intervals and transformed into a one-hot encoded vector.

\subsubsection{In-Degree and Out-Degree Histograms}
These histograms represent the distribution of connections that go into and exit nodes, providing information on the directional flow of the graph. Each histogram uses different calculations.
\begin{enumerate}
    \item[a.] \textbf{In-degree} is calculated by counting the incoming edges to each node and categorizing these counts into bins. The counts are then compiled into a histogram.
    \item[b.] \textbf{Out-degree} similarly counts and categorizes the outgoing edges from each node into a histogram.
\end{enumerate}
}
\subsection{Image Features}
\label{appendix:image-features}
Image features are extracted using EfficientViT. The images are resized, normalized, and then processed to obtain average-pooled features. Dimensionality reduction is applied using Principal Component Analysis (PCA), reducing the feature space to the top components that capture the most variance.

\subsection{Text Features}
Text data from prompts is processed using a TF-IDF vectorizer followed by a Truncated Singular Value Decomposition (SVD) to reduce the dimensionality while capturing latent semantic information. The number of components retained is based on the model's explanatory power on the variance observed in the text data.

\subsection{Interaction Features}
Interaction terms are created to model the interaction between different types of features, such as the interaction between the number of nodes and the width of the edges. These features are expected to provide insights into how structural properties might influence the overall characteristics of the graphs.

\subsection{Edge Width Features}
The set of possible edge width parameter values are converted into one-hot encoded vectors.

\subsection{Node Color Features}
Each channel of a node color's RGB values are split and added as continuous features.

\subsection{Encoding and Normalization}
Categorical variables such as graph types and task identifiers are one-hot encoded. Continuous features are normalized or standardized to ensure they contribute equally to the model training process.

\subsection{Feature Reduction}
Post-feature extraction and engineering, features undergo a selection process in which constant features are removed and the remaining features are used for model training. This step ensures that only the most informative features are considered, improving model performance and computational efficiency.

\subsection{Matrix Construction}
Finally, all engineered features are combined into a single feature matrix, along with target labels for classification tasks. This matrix serves as an input for the subsequent model training phase.

\section{Classification}
\label{appendix:classification}

The classification framework is trained to predict ground truth values from engineered features derived from text-image paired prompts. The following classification metrics show the performance of the logistic regression models:

\begin{table}[htbp]
    \centering
    \resizebox{\columnwidth}{!}{%
    \begin{tabularx}{\columnwidth}{Xcccc}
    \toprule
    Model & Accuracy & Precision & Recall & F1 \\
    \midrule
    GPT-4o & 0.828 & 0.838 & 0.849 & 0.843 \\
    GPT-4 Vision Preview & 0.845 & 0.838 & 0.875 & 0.856 \\
    Gemini 1.5 Pro & 0.840 & 0.808 & 0.837 & 0.822 \\
    Gemini 1.5 Flash & 0.866 & 0.872 & 0.903 & 0.887 \\
    Gemini 1.0 Pro Vision & 0.845 & 0.815 & 0.837 & 0.826 \\
    Claude 3 Opus & 0.838 & 0.781 & 0.629 & 0.697 \\
    Claude 3 Sonnet & 0.823 & 0.755 & 0.668 & 0.709 \\
    Claude 3 Haiku & 0.814 & 0.713 & 0.673 & 0.693 \\
    \bottomrule
    \end{tabularx}%
    }
    \caption{The classification model performances on each LMM's evaluation.}
    \label{table:classification}
\end{table}

\begin{figure*}[htbp]
    \centering
    \includegraphics[width=\textwidth]{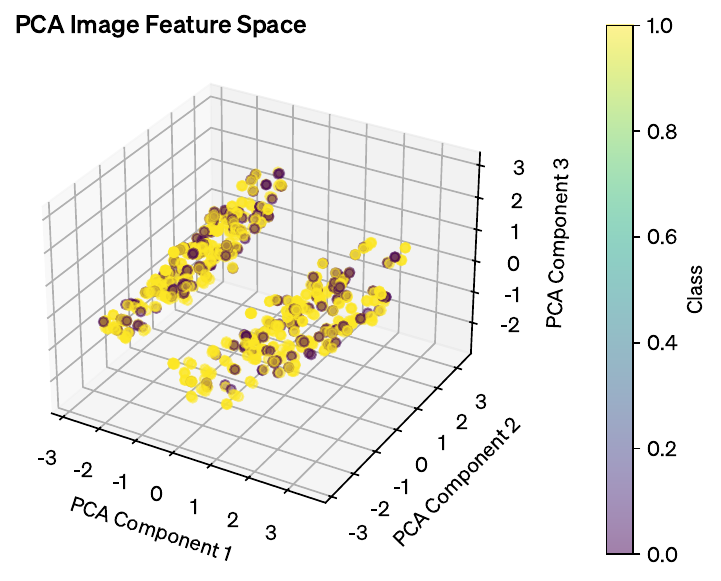}
    \caption{A 3D perspective of the PCA image feature space from the first three components.}
    \label{fig:pca-3d}
\end{figure*}

\begin{figure*}[htbp]
    \centering
    \includegraphics[width=\textwidth]{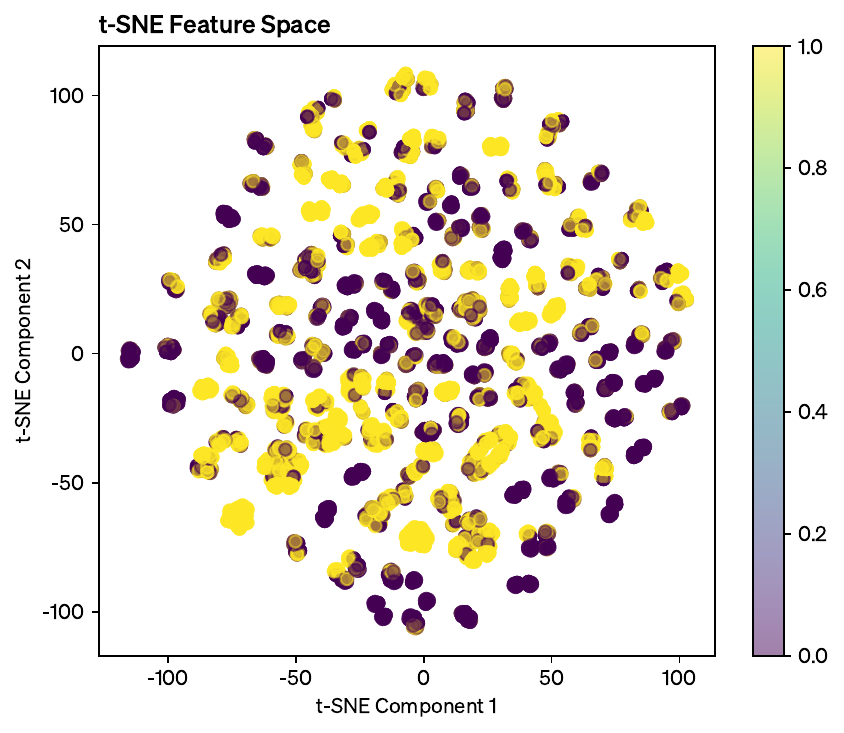}
    \caption{The t-SNE feature space from the first two components.}
    \label{fig:tsne-2d}
\end{figure*}

\end{document}